# Forecasting Electric Vehicle Battery Output Voltage: A Predictive Modeling Approach

Narayana Darapaneni
*Director - AIML*
Great Learning
Bengaluru, India
darapaneni@gmail.com

Ashish K
*Student - DSML*
PES-Great Learning
Bengaluru, India
ashishon26@gmail.com

Ullas M S
*Project Mentor*
PES-Great Learning
Bengaluru, India
rao.ullas@gmail.com

Anwesh Reddy Paduri
*Senior Data Scientist*
PES-Great Learning
Bengaluru, India
anwesh@greatlearning.in

*Abstract* — The battery management system plays a vital role in ensuring the safety and dependability of electric and hybrid vehicles. It is responsible for various functions, including state evaluation, monitoring, charge control, and cell balancing, all integrated within the BMS. Nonetheless, due to the uncertainties surrounding battery performance, implementing these functionalities poses significant challenges. In this study, we explore the latest approaches for assessing battery states, highlight notable advancements in battery management systems (BMS), address existing issues with current BMS technology, and put forth possible solutions for predicting battery charging voltage.

## I. Introduction and Literature Review

The rapid growth in the adoption of electric vehicles (EVs) in recent years has emphasized the need for efficient battery management systems (BMS) that can accurately predict charging voltage. The BMS plays a critical role in maintaining optimal battery performance, extending battery life, and improving the overall efficiency of EVs. By accurately forecasting the charging voltage, the BMS can optimize the charging process, prevent overcharging or undercharging, and ensure the longevity of the battery.

Traditionally, BMSs have relied on rule-based algorithms or simplistic models to estimate the charging voltage. However, these approaches often struggle to capture the complex interactions between various battery parameters and charging voltage accurately. As a result, the accuracy of voltage predictions may be limited, leading to suboptimal charging strategies and potential damage to the battery. To address these challenges, this paper proposes a novel approach that leverages machine learning techniques to predict charging voltage in EVs.

Machine learning has proven to be highly effective in analyzing complex and large datasets, making it an ideal tool for capturing the intricate relationships between battery characteristics and charging voltage. This research helps us to understand, the scope of ML in the applications of electric vehicle battery management system, leveraging ML model capabilities to improve the accuracy and efficacy of battery management systems for electric vehicles.

This study underscores the significance of achieving optimal performance and longevity of electric vehicle (EV) batteries through the precise prediction of their charging voltage. By doing so, the electric vehicle battery management system (BMS) can be employed to its fullest potential.
Accurate forecasting of the charging voltage plays a pivotal role in maintaining the battery in an ideal operational state. It ensures that the battery operates within its designated voltage range, preventing overcharging or undercharging scenarios. These extremes can be detrimental to the battery's health, causing irreversible damage and potentially reducing its lifespan.
Furthermore, this predictive capability contributes to the overall enhancement of the efficiency and effectiveness of the battery management system. By consistently monitoring and regulating the charging voltage in line with anticipated requirements, the BMS can proactively manage the battery's state of charge (SOC) and state of health (SOH). This proactive management allows for optimal energy utilization, as the BMS can adjust charging and discharging cycles to align with the battery's predicted behavior.
In essence, the research emphasizes that accurate charging voltage prediction is a linchpin for achieving several critical objectives within the realm of EV battery management. It ensures battery health, maximizes efficiency, and empowers the BMS to fulfill its role in maintaining and optimizing the battery's performance.

The battery management system (BMS) is a critical component in ensuring the safety and reliability of electric and hybrid vehicles by monitoring and controlling parameters such as battery output voltage, current, state assessment, charge regulation, and cell balancing. However, the inherent uncertainty in battery performance poses significant challenges in the effective execution of these essential BMS functions. Of particular concern is the accurate prediction of battery charging voltage, which plays a pivotal role in optimizing vehicle performance, energy efficiency, and safety.

Addressing these challenges is imperative for advancing BMS technology in electric and hybrid vehicles. Current methodologies for evaluating battery output voltage within BMS solutions often fall short in capturing the intricate nuances of battery behavior, leading to suboptimal real-time predictions. Consequently, this limitation impacts vehicle performance, energy management decisions, and overall operational efficiency.

The inability to precisely predict battery charging voltage not only hampers power optimization but also contributes to range anxiety and erodes confidence in the reliability of electric vehicles. To overcome these obstacles, there is a pressing need for innovative approaches and technologies that can enhance the accuracy of battery charging voltage predictions within BMS systems, thus driving the progress and broader acceptance of electric and hybrid vehicles.

Through the successful resolution of the core challenge of predicting battery charging voltage, this study will bring about a substantial enhancement in the capabilities of battery management systems (BMS) employed in electric and hybrid vehicles. The proposed advancements in predictive modeling hold the potential to revolutionize range estimation, power management optimization, and safety enhancements, thereby fostering the broader acceptance of electric and hybrid vehicles and contributing to a sustainable and environmentally friendly future for transportation.

The growing popularity of electric vehicles (EVs) as an eco-conscious mode of transport highlights the pivotal role of the battery management system (BMS). The BMS assumes responsibility for the vigilant monitoring and regulation of critical parameters such as temperature, state of charge (SOC), and state of health (SOH) within the battery pack. An effectively designed and finely-tuned BMS can yield substantial improvements in battery performance, safety, longevity, and, by extension, the overall driving range and efficiency of EVs.

Consequently, the research interest in the development of sophisticated BMS solutions for EVs has surged. This comprehensive literature review endeavors to offer a meticulous overview of the recent advancements and accomplishments in the realm of BMS for EVs, with a particular emphasis on the model predictive control (MPC) technique, state-of-charge estimation, and thermal management. Notably, the literature reveals a discernible upward trajectory in the utilization of BMS for orchestrating energy storage systems in EVs, a trend driven by the potential to optimize battery performance and prolong battery life. The focal points of this review encompass the following key domains:

A. *State-of-Charge Estimation*

State of Charge (SOC) estimation stands as a pivotal component within battery management systems, serving as a barometer of a battery's remaining energy, expressed as a fraction of its total capacity. Accurate SOC estimation plays an indispensable role in optimizing battery performance, predicting battery life, and safeguarding against overcharging and overdischarging.

Several techniques are employed for SOC estimation, including Coulomb counting, open-circuit voltage (OCV) measurement, and Kalman filtering. Coulomb counting quantifies the charge entering or departing the battery over time to compute SOC, whereas OCV measurement involves evaluating the battery's voltage against a lookup table of voltage-SOC relationships. Kalman filtering, a more intricate method, derives SOC by considering current and voltage measurements through a mathematical battery behavior model.

To attain the highest precision, a blend of these methods is often preferred, given that each approach has its distinct merits and limitations. SOC estimation constitutes an active field of research, continually advancing to develop more accurate and reliable methodologies adaptable to various battery chemistries and applications.

In a comprehensive analysis, Uddin et al. [7] scrutinize state-of-the-art SOC estimation techniques for EV batteries, encompassing voltage-based, current integration-based, Kalman filter-based, adaptive observer-based, and neural network-based methodologies. Challenges and constraints of each approach are delineated, offering valuable insights for future research directions.Li et al. [8] propose a neural network-based SOC estimation method for EV batteries, leveraging data from battery voltage, current, and temperature. Simulation outcomes substantiate the method's high accuracy and resilience across diverse operational conditions.

Reference [9] provides an exhaustive review of various SOC estimation techniques, encompassing Kalman filters, neural networks, and adaptive observers. The authors elucidate the merits and demerits of each method, rendering guidance for selecting the most apt SOC estimation technique contingent upon distinct EV applications.

B. *Battery Thermal Management*

Within the realm of battery management systems (BMS), the importance of battery thermal management (BTM) looms large as a cornerstone for ensuring efficient and safe battery operation. BTM achieves this by vigilantly monitoring and adjusting the temperature of individual battery cells within a pack, thereby maintaining the battery pack's thermal equilibrium.

Several methodologies are employed for battery thermal management, encompassing both active and passive approaches. Active cooling leverages cooling fluids, such as liquid or air, to dissipate heat from the battery pack. Conversely, passive cooling capitalizes on materials with high thermal conductivity to naturally redirect heat away from the battery pack.

One of the prevailing techniques for battery thermal management is active liquid cooling. In this method, a coolant fluid, typically water or glycol, circulates through a network of tubes or channels within the battery pack. The fluid absorbs heat emanating from the battery cells and conveys it to a heat exchanger, where it is released into the ambient environment. This approach ensures precise temperature regulation and remains widely adopted.

Thermal management bears paramount significance for the dependable and secure operation of batteries. Low temperatures can compromise battery performance and efficiency, while elevated temperatures accelerate degradation, curtailing battery lifespan. Furthermore, overheating can trigger thermal runaway, posing the risk of fires or explosions.

Duan et al. [11] undertake an exhaustive exploration of contemporary advancements in battery thermal management systems (BTMS) for electric vehicles (EVs). Their investigation encompasses active cooling, passive cooling, and phase-change cooling techniques. While elucidating the merits and drawbacks of each approach, the authors identify research challenges and prospects.

Reference [3] proposes the application of Model Predictive Control (MPC) to govern the thermal performance of lithium-ion batteries in electric vehicles. This method hinges on battery temperature and state of charge (SOC) estimation, coupled with a thermal model of the battery. The authors substantiate the efficacy of their approach in heightening battery performance and extending battery life.

In a comprehensive review, [12] delves into the realm of battery thermal management (BTM) systems deployed in electric vehicles. The authors dissect active, passive, and hybrid systems, probing into the unique strengths and limitations of each strategy. Active systems modulate battery pack temperature through refrigeration cycles or heat pumps, while passive systems harness high thermal conductivity materials for heat dissipation. Hybrid systems, combining both approaches, aspire for optimal temperature control. Various cooling techniques, including liquid and air cooling, are also scrutinized.

The paper emphasizes the pivotal role of battery thermal management in electric vehicles, impacting battery pack performance, longevity, and safety. It underscores the multifaceted factors such as battery chemistry, size, and shape that wield influence over thermal management. Moreover, it accentuates the imperative integration of the BTM system with the vehicle's climate control infrastructure.

Reference [13] reiterates the significance of BTM in electric vehicles, elucidating its ramifications on battery pack performance, lifespan, and safety. The authors navigate through diverse BTM strategies, including active, passive, and hybrid systems, coupled with an exploration of cooling techniques such as liquid and air cooling. To address challenges, the authors proffer solutions, including the adoption of cutting-edge materials and optimization of BTM systems.

In [14], the authors present a comprehensive design and analysis of a battery thermal management (BTM) system tailor-made for electric vehicles. Their innovative BTM system incorporates a thermal management module and a liquid cooling system. The thermal management module capitalizes on phase-change materials (PCM) to absorb and release heat during charge and discharge cycles. Simulation outcomes affirm the system's prowess in regulating the battery pack's temperature with a temperature variance of less than 3°C. The adoption of phase-change materials (PCM) in the thermal management module, recognized for their superior heat storage capacity and cost-efficiency, is a noteworthy facet.

Collectively, these references demystify the realm of Battery Thermal Management (BTM) and underscore its pivotal role in the context of electric vehicles. They explore an array of BTM strategies, including active, passive, and hybrid systems, unveiling the nuanced advantages and limitations inherent to each approach. Additionally, these references delve into various cooling techniques, spanning from liquid to air cooling, collectively presenting a holistic overview of BTM in the electric vehicle landscape.

## C. Energy Management

The Energy management encompasses the processes of monitoring, regulating, and conserving energy within a structure or system. Its primary objective is to minimize energy wastage while maximizing efficient energy utilization. Achieving this goal involves the implementation of various energy-saving measures, such as the incorporation of renewable energy sources, energy-efficient equipment, and building automation systems. This article delves deeply into the concept of energy management, shedding light on key tools and methodologies integral to its implementation.

Load forecasting stands out as a pivotal technique within energy management. It entails the prediction of future energy requirements for a building or system. This prediction is based on a comprehensive analysis of historical energy consumption data, weather-related information, and other pertinent variables. Subsequently, predictive models are constructed using this data to anticipate energy usage accurately. By dynamically adjusting energy consumption in response to expected demand, load forecasting empowers building managers to optimize energy utilization and curtail wastage effectively.

The integration of renewable energy sources constitutes another cornerstone of energy management. Renewable sources like solar and wind energy offer sustainable and environmentally friendly energy solutions. These sources can be seamlessly integrated into a building's energy infrastructure through the deployment of technologies such as solar panels, wind turbines, and other renewable energy systems. Leveraging renewable energy mitigates reliance on fossil fuels, reducing the carbon footprint of buildings and systems.

Building automation systems emerge as indispensable tools in the realm of energy management. These systems are adept at monitoring and controlling various facets of a building's energy consumption, including lighting, heating, and cooling. Through automation, building managers can fine-tune energy utilization, minimizing waste. For instance, automated lighting systems can intelligently switch off lights in unoccupied areas, while automated heating and cooling systems can adjust temperature settings based on occupancy and environmental conditions.

Energy storage technology assumes a pivotal role in effective energy management. Devices such as batteries and flywheels serve as energy reservoirs, capable of storing surplus energy generated from renewable sources for future use. This approach minimizes energy wastage and ensures a consistent energy supply during peak demand periods. Additionally, energy storage systems can function as reliable backup power sources during unforeseen power outages.

In [15], the authors provide an in-depth overview of lithium-ion battery energy management, pinpointing the limitations of existing energy management strategies, including the need for more precise battery diagnosis and prognosis. To enhance the accuracy and reliability of energy management in electric vehicles (EVs), the authors recommend the integration of diagnostic and prognostic techniques rooted in machine learning algorithms.

Reference [1] explores the application of model predictive control (MPC) within building and energy management systems. The paper underscores MPC's ability to optimize energy production and consumption while adhering to control constraints. It meticulously examines the application of MPC across various building energy systems, spanning lighting, heating, ventilation, and air conditioning.

Wen et al. [16] propose a real-time energy management system for electric buses, grounded in a battery model. This system leverages the battery model to forecast the battery's state of charge and the bus's energy consumption, subsequently optimizing battery charging and discharging to curtail energy consumption.

## D. Design of BMS

A Battery Management System (BMS) plays a pivotal role in monitoring and ensuring the safe and efficient operation of a battery pack, a critical component in electric vehicles and other battery-driven applications. Designing an effective BMS necessitates meticulous consideration of several crucial elements and factors:

***Cell Balancing***: A primary responsibility of the BMS is to maintain cell balance within the battery pack, ensuring consistent state of charge and state of health across all cells. Effective cell balancing can be achieved through either passive or active balancing methods.
Protection Circuitry: The BMS must safeguard the battery pack from overcharging, over-discharging, and extreme temperature conditions. This involves incorporating protection circuitry like overvoltage and undervoltage protection, along with temperature sensors.

***Communication***: Communication capabilities are essential, enabling the BMS to interact with other systems such as the vehicle or charger. Compatibility with communication protocols like CAN or LIN facilitates data transmission and reception.

***Current and Voltage Sensing***: Precise measurement of current and voltage in each battery cell is fundamental for calculating state of charge (SOC), state of health (SOH), and temperature.

***Software***: BMS software plays a critical role in observing and managing battery pack operation. It should align with hardware and communication requirements.

***Redundancy***: To ensure continued operation in the face of component failures, the BMS should incorporate redundancy.

***Scalability***: Designing the BMS to be scalable enables it to adapt to various battery pack sizes and chemistries.

Chen et al. [17] proposed an innovative multi-objective optimization-based design approach for BMS. Their method optimizes cost, dependability, and safety aspects of BMS design by utilizing a comprehensive battery model accounting for aging and temperature effects. Oliveira et al. [18] delved into challenges surrounding battery management systems for electric vehicles, addressing issues like battery degradation, charging infrastructure, and safety, offering insights for future research. Nguyen et al. [19] provided a comprehensive review of next-generation green vehicles, including battery-powered electric vehicles, hybrids, and fuel cells, evaluating technologies, advantages, and disadvantages.

Viswanathan et al. [20] conducted a comparative analysis of various BMS architectures for electric and hybrid electric vehicles, considering capabilities, complexity, and costs. They assessed centralized, distributed, and modular architectures while discussing their merits and drawbacks. Zhang et al. (2019) [22] introduced a BMS design based on an embedded system featuring a microcontroller unit (MCU) and peripheral devices. They detailed the hardware and software architecture and validated its effectiveness through experiments. [23] presented a novel modular-based approach to cell balancing in lithium-ion battery packs, proposing a control algorithm for low-voltage level cell balancing, substantiating its efficacy through simulations and experiments.

Paper [24] introduced an adaptive extended Kalman filter (AEKF) for lithium-ion battery state estimation, demonstrating its effectiveness through simulations and experiments, comparing it to alternative state estimation methods.[25] detailed the development of a wireless BMS for electric vehicles (EVs), describing the hardware and software architecture and validating its performance through experiments.[26] conducted an extensive analysis of state of charge (SOC) estimation techniques for lithium-ion batteries in electric vehicles, comparing and evaluating methods like model-based approaches, Kalman filters, and neural networks.Article [27] proposed an optimal charging and discharging control algorithm for EVs founded on a BMS, demonstrating its effectiveness through simulations and experiments, and comparing it to alternative control strategies introduced a low-resistance resistor-based cell balancing control strategy for lithium-ion battery packs, proposing a control algorithm based on voltage, and substantiating its efficacy through simulations and experiments.

E. *Fault Diagnosis and Prognosis*

Fault diagnosis and prognosis represent pivotal facets of any system, including Battery Management Systems (BMS), playing a crucial role in the timely identification and anticipation of potential issues within the battery system to facilitate timely maintenance and repair.This report delves into diverse fault diagnosis and prognosis techniques employed in BMS:

Deep Learning-Based Fault Diagnosis and Prognosis - [29] introduced a deep learning-based method capable of fault identification, resolution, and the prediction of the battery's remaining useful life.

Model-Based Fault Diagnosis - Model-based approaches involve crafting a mathematical representation of the battery system, contrasting the actual system's behavior with model predictions. In [30], various model-based techniques for fault diagnosis and prediction are discussed. Widely used methods include Extended Kalman Filter (EKF), Particle Filter (PF), and Unscented Kalman Filter (UKF), leveraging parameters such as State of Charge (SOC), State of Health (SOH), and others to identify faults and anticipate future behavior.

Data-Driven Fault Diagnosis - Data-driven approaches circumvent the need for a mathematical model by analyzing extensive data from the battery system via machine learning techniques. [31] employs an Artificial Neural Network (ANN) to diagnose faults, trained on data from both healthy and faulty batteries to identify anomalies and predict future faults.

Signal Processing-Based Fault Diagnosis - Signal processing techniques analyze signals from the battery system, such as voltage and current readings. In [32], wavelet transform and Support Vector Machine (SVM) are combined to identify battery system issues. Wavelet transform extracts signal features, which are then input into the SVM classifier for problem detection.

Prognosis Techniques - Prognosis techniques predict the battery system's future behavior, enabling proactive maintenance and repair. In [33], Gaussian Mixture Model (GMM) and Support Vector Regression (SVR) are used to estimate the battery's Remaining Useful Life (RUL). GMM categorizes data into distinct groups based on similarity, and SVR predicts RUL for each group.

Combined Approaches - Researchers have proposed combining various fault diagnosis and prognosis techniques to enhance accuracy and reliability. [34] employs a combination of wavelet transform, extreme learning machine, and particle swarm optimization (PSO) to identify battery system issues. PSO optimizes feature extraction, while ELM serves as a classifier.

In conclusion, fault diagnosis and prognosis are integral components of Battery Management Systems (BMS). An array of techniques, including model-based, data-driven, signal processing-based, and prognosis methods, are leveraged to identify faults and anticipate future behavior. The fusion of diverse techniques can augment accuracy and dependability, with future research poised to yield even more sophisticated and efficient methodologies.

II. ARCHITECTURE DIAGRAM

The proposed methodology involves two stages. First stage, is where a base model is build with various models to analyze the data and understand the data set. Second stage involves extensive feature elimination and optimization over the base model.

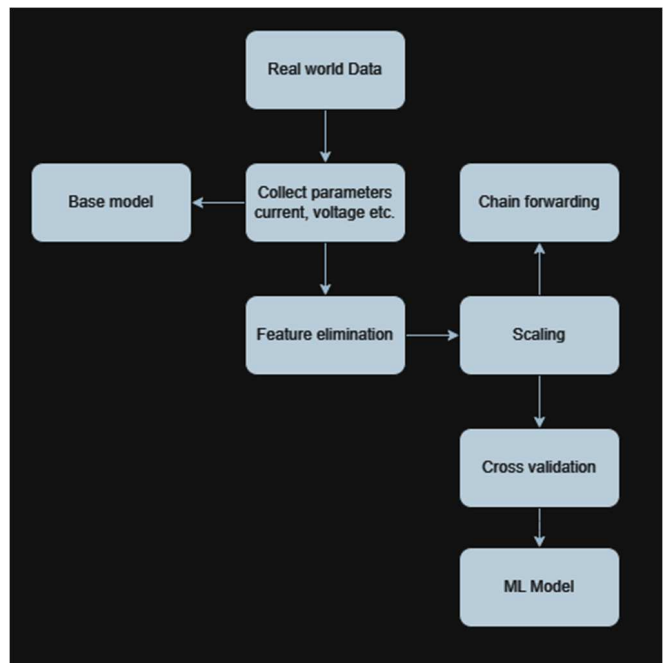

Figure 1: Architecture Diagram of BMS

A. *Data Pre-processing*

The historical dataset pertaining to battery charging, encompassing additional variables such as temperature, Ecell voltage, current, and cycle number, undergoes a preprocessing phase. During this process, measures are taken to address missing values, and the data is standardized to ensure uniform scaling.

B. *LSTM-Network*

The processed dataset is subjected to LSTM modeling, illuminating concealed patterns within the dataset and affording us insights into its structure while facilitating the analysis of pivotal data points.

C. *Training and Testing*

Following the training of the LSTM network on the pre-processed battery discharge data, the network becomes adept at predicting the battery's charging voltage by assimilating input sequences of measurements. The LSTM model is then executed on the training dataset, where performance metrics are meticulously assessed. Subsequently, the model is put through its paces on a separate test dataset to gauge its overall performance during a later stage of evaluation.

D. *Performance Evaluation*

The LSTM network's predictions are subsequently scrutinized alongside those generated by widely employed predictive models, including the linear regressor, random forest, decision tree, KNN, and SGD. This evaluation encompasses an array of performance metrics, including but not limited to RMSE, mean-absolute-percentage-error (MAPE), mean-absolute-error (MAE), and mean-squared-error (MSE). This comparative analysis provides valuable insights into the model's accuracy and effectiveness, enabling a comprehensive assessment of its performance.

E. *Model Prediction Framework*

To determine the effective charging voltage of the battery, a base model is built with LSTM network to reveal complex patterns in the data set, and in the later stages of the framework, various parameters are tuned and LSTM is carried out in the preprocessed data set.

The model framework comprises primarily of a two-stage process. In the initial stage, the foundation is laid by constructing a fundamental model employing well-established predictive algorithms commonly employed in the field. Subsequently, during the second stage, the model is further refined and enhanced by meticulously fine-tuning its parameters. This iterative phase is undertaken with the express purpose of optimizing and refining the model, elevating its predictive capabilities to a higher level of performance.

Feature Extraction: This pertains to the procedure of transmogrifying unprocessed data into a compendium of pertinent attributes meticulously designed to encapsulate quintessential information. This intricate process encompasses methodologies ranging from the reduction of dimensionality and meticulous filtration to the judicious application of mathematical metamorphosis, all orchestrated with the singular objective of distilling the most perspicacious characteristics from the dataset. In our particular context, these characteristics encompass those governing both the entrustment and discharge of responsibilities. The overarching aim is to mitigate data intricacy, expunge superfluous or extraneous particulars, and accentuate salient patterns or attributes that are pivotal for subsequent scrutiny or model development.

Feature Fusion: This process, alternatively referred to as feature amalgamation or feature amalgamation, entails the amalgamation or integration of numerous features into a unified composite feature. This amalgamation is undertaken to apprehend complementary information originating from diverse sources, thereby enhancing the representational potency and discriminatory efficacy of these features. In the context of our dataset, we employ an Auto-Encoder to execute this operation. Such utilization fortifies the data representation by capitalizing on the distinct strengths emanating from various feature sources or modalities.

Charging voltage Prediction: The prediction of battery charging voltage holds paramount significance across numerous domains, encompassing asset management, predictive maintenance, and the realm of reliability engineering. This task is centered on the estimation of remaining battery lifespan or the time until failure of a system, component, or asset, grounded in the analysis of available data and predictive models. Models for battery charging voltage prediction rely on historical data, sensor measurements, and pertinent features to anticipate the forthcoming degradation and potential system failure. This predictive capability fosters proactive decision-making, facilitates the optimization of maintenance schedules, minimizes operational downtime, and maximizes the efficiency and utilization of assets. The pursuit of battery charging voltage prediction may encompass diverse methodologies, including statistical approaches, machine learning algorithms, and time series analysis, contingent upon the data's nature and the specific requisites of the problem at hand. In our research, we adopt a Long Short-Term Memory (LSTM) network applied to standardized data as the chosen approach for predicting battery charging voltage.

III. OPTIMIZATION OF THE MODEL

Upon establishing these foundational models, we implement K-fold cross-validation. This technique entails dividing the provided dataset into K equitably sized subsets or folds and is employed to evaluate the efficacy of machine learning models. The process encompasses iteratively training and testing the model K times, employing the first K-1 folds for training and the remaining fold as the testing set. Utilizing K-fold cross-validation enables a more comprehensive evaluation of the base models' performance, offering a superior insight into how the model is expected to perform on unseen data compared to a single train-test split.

K-fold cross-validation enhances the dependability of the evaluation process since it accounts for various permutations of training and testing datasets. This approach mitigates the potential bias stemming from a sole train-test division, resulting in a more precise evaluation of the model's ability to generalize to unobserved data. A pronounced enhancement in the performance of the LSTM model is readily discernible, while noteworthy improvements have also been observed in several other models. Furthermore, we have embarked on additional optimization endeavors by systematically adjusting various model parameters to further elevate the model's efficacy. Notably, measures have been taken to address overfitting concerns, resulting in the mitigation of such issues in certain models. Parameters of significance, including epoch and batch size, have been meticulously fine-tuned to harmonize with the dataset's characteristics.

IV. RESULTS AND MODELS COMPARISON

.To assess the model's accuracy, we have employed a range of evaluation metrics, including:

Mean Absolute Error (MAE): MAE, a straightforward regression metric, measures the average absolute difference between predicted and actual values. Unlike MSE and RMSE, MAE does not involve squaring the errors, rendering it less sensitive to outliers. This characteristic enhances its robustness when assessing model performance with datasets containing extreme values. A lower MAE signifies higher predictive accuracy. MAE is especially valuable when seeking a metric that characterizes typical or median prediction errors without being disproportionately affected by a few extreme errors. It offers an intuitive understanding of the model's average prediction accuracy within the context of the problem at hand.

Root Mean Squared Error (RMSE): RMSE, a derivative of MSE, calculates the square root of the average squared disparities between predicted and actual values. RMSE is particularly advantageous when a metric is needed in the same units as the target variable. By employing the square root, RMSE transforms squared errors back to the original scale, enhancing interpretability. Similar to MSE, RMSE assigns heavier penalties to larger errors, offering insight into the average magnitude of prediction errors. A lower RMSE denotes a better model fit to the data. It is commonly used when precision is required to align with the units of the problem domain.

Mean Squared Error (MSE): MSE, a foundational regression metric, serves as a measure for evaluating predictive model performance. It quantifies the average squared difference between model-generated predictions and actual observed values within the dataset. In essence, MSE gauges the variance or dispersion of prediction errors. Through the squaring of errors, MSE penalizes larger errors more significantly than smaller ones. This characteristic can render it sensitive to outliers. A lower MSE indicates closer alignment between the model's predictions and actual values, implying a superior fit to the data. Due to its mathematical tractability and capability to capture both systematic and random errors, MSE enjoys widespread use in regression tasks. Nevertheless, it is essential to interpret MSE within the specific problem context, considering its scale depends on the target variable's units.

Mean Absolute Percentage Error (MAPE): Mean Absolute Percentage Error (MAPE) is a metric employed to evaluate prediction accuracy in regression tasks, particularly when assessing the relative magnitude of errors in comparison to actual values. MAPE calculates the average percentage difference between predicted and actual values, expressing errors as percentages of actual values. A lower MAPE signifies heightened prediction accuracy. MAPE proves valuable when assessing the relative impact of errors across diverse data points, especially when dealing with data featuring varying scales or magnitudes.

### A. Base Model Comparison

The table presented below illustrates the performance metrics of diverse predictive models. The findings suggest that, in this particular context, LSTM exhibits relatively poorer performance. It's important to note that this analysis represents an initial assessment, serving as a foundation for comprehending the data and guiding subsequent enhancements. .

Table 1: Evaluation result of base model

| Model | MSE | RMSE | MAE |
|---|---|---|---|
| Linear regressor | 1.77 | 1.33 | 1.19 |
| Neural Network | 3.91 | 1.97 | 1.64 |
| LSTM | 39.94 | 6.32 | 6.18 |
| SGD | 1.24 | 1.11 | 9.79 |
| Random Forest | .54 | .73 | .52 |
| Decision Tree | .63 | .79 | .78 |
| KNN | 1.98 | 1.40 | .96 |

From the aforementioned outcomes, we can draw the following conclusions: Random forest and decision tree models exhibit the lowest RMSE values, signifying superior predictive accuracy compared to other models. However, it's essential to consider the possibility of overfitting in these models due to their exceptionally low RMSE values. On the contrary, LSTM and Neural Networks demonstrate the highest RMSE values, suggesting that they may not be well-suited for this particular dataset. Fine-tuning the model parameters could potentially enhance their performance on this dataset.

### B. Performance analysis of Models with K-Fold Cross Validation

Upon establishing these foundational models, we implement K-fold cross-validation. This technique entails dividing the provided dataset into K equitably sized subsets or folds and is employed to evaluate the efficacy of machine learning models. The process encompasses iteratively training and testing the model K times, employing the first K-1 folds for training and the remaining fold as the testing set. Utilizing K-fold cross-validation enables a more comprehensive evaluation of the base models' performance, offering a superior insight into how the model is expected to perform on unseen data compared to a single train-test split.

Table 2: Models with K-Fold Cross Validation

| Model | Mean RMSE | MSE |
|---|---|---|
| Linear regressor | 1.03 | 1.06 |
| Decision Tree | 1.13 | 1.27 |

In the presented findings, it is evident that the utilization of k-fold cross-validation enhances the dependability of the evaluation process since it accounts for various permutations of training and testing datasets. This approach mitigates the potential bias stemming from a sole train-test division, resulting in a more precise evaluation of the model's ability to generalize to unobserved data.

Analysis of the cross-validation scores indicates notable enhancements in the performance of the Linear Regressor, particularly in terms of the RMSE score. Additionally, it appears that overfitting, which was previously observed in the Decision Tree model, has been partially mitigated. Further refinements and optimization efforts will be undertaken to further enhance the model's performance.

### C. Evaluation of Models after Optimization

The following table presents the outcomes achieved subsequent to the implementation of optimization procedures within the models. A pronounced enhancement in the performance of the LSTM model is readily discernible, while noteworthy improvements have also been observed in several other models. Furthermore, we have embarked on additional optimization endeavors by systematically adjusting various model parameters to further elevate the model's efficacy. Notably, measures have been taken to address overfitting concerns, resulting in the mitigation of such issues in certain models. Parameters of significance, including epoch and batch size, have been meticulously fine-tuned to harmonize with the dataset's characteristics. The ensuing table provides an illustrative summary of the results following the parameter tuning process.

Table 3: Optimizes models Results

| Model | MSE | RMSE | MAE | MAPE |
|---|---|---|---|---|
| Linear Regressor | 1.04 | 1.09(mean) | | |
| LSTM Stage1 | 2.45 | 1.56 | 1.04 | .52 |
| LSTM Stage2 | 1.95 | 1.39 | .92 | .13 |
| Random Forest | .94 | .97 | .48 | .06 |

Subsequent refinements to the model have yielded substantial enhancements, culminating in improved predictive capabilities. In the initial stage of LSTM modeling, we initiated modifications to critical parameters like epoch and batch size. Subsequently, we delved into a more extensive fine-tuning process during the second stage of LSTM modeling. These deliberate adjustments and optimizations have notably bolstered the model's predictive prowess. Additionally, it's noteworthy that the Linear Regressor model has also exhibited considerable advancement in performance during this optimization stage. These refinements collectively signify the efficacy of the optimization process in enhancing model performance and predictive accuracy.

D. *Comparative Analysis of LSTM at each stage*

The following table provides a comprehensive overview of the performance metrics for the LSTM model at each stage of its development. It is distinctly observable that the model's performance experiences substantial enhancement as it progresses through the stages. A meticulous approach encompassing Exploratory Data Analysis (EDA), judicious feature elimination, and precise parameter tuning is rigorously executed at each juncture, consequently contributing to the model's increasingly accurate predictions.

Numerous performance metrics, including Root Mean Squared Error (RMSE), Mean Absolute Percentage Error (MAPE), Mean Absolute Error (MAE), and Mean Squared Error (MSE), exhibit noteworthy iterative improvements. These advancements can be attributed to a multitude of factors, prominently feature elimination, the standardization of data, the systematic elimination of outliers, and the fine-tuning of crucial model parameters. Cumulatively, these refinements and optimizations represent a significant stride toward enhancing the model's predictive capacity and overall performance

Table 4: Comparative results of LSTM stages

| Model | MSE | RMSE | MAE | MAPE |
|---|---|---|---|---|
| LSTM Base model | 39.94 | 6.32 | 6.18 | 5.84 |
| LSTM Stage1 | 2.45 | 1.56 | 1.04 | .52 |
| LSTM Stage2 | 1.95 | 1.39 | .92 | .13 |

## V. CONCLUSION

In conclusion, the LSTM model's performance has exhibited significant enhancements at each stage of its development. This progress can be attributed to a systematic and meticulous approach that encompassed Exploratory Data Analysis (EDA), feature elimination, and precise parameter tuning. As the model evolved, it became increasingly capable of making accurate predictions.

The improvements are evident across various performance metrics, such as Root Mean Squared Error (RMSE), Mean Absolute Percentage Error (MAPE), Mean Absolute Error (MAE), and Mean Squared Error (MSE). These metrics demonstrated substantial iterative enhancements, underlining the effectiveness of the applied strategies.

Key factors contributing to the model's enhanced performance include feature elimination to reduce noise and redundancy, standardization of data for improved consistency, systematic outlier elimination to ensure data integrity, and fine-tuning of critical model parameters. The cumulative impact of these refinements and optimizations signifies a noteworthy achievement in augmenting the LSTM model's predictive capabilities and overall performance.

This comprehensive analysis underscores the significance of a structured approach to model development, highlighting how systematic data preprocessing, feature engineering, and parameter tuning can significantly enhance the predictive accuracy of machine learning models. These findings can serve as valuable insights for future model development and optimization efforts, emphasizing the importance of iterative refinement and a data-driven approach to achieve superior model performance..

## VI. FUTURE SCOPE

Future research in this area offers several promising directions for further advancement. First and foremost, there is room for additional optimization of predictive models. This includes exploring advanced parameter tuning techniques, feature engineering methods, and data preprocessing strategies to enhance model performance. Ensemble learning techniques, which combine the strengths of multiple models, could also be investigated to achieve even more accurate predictions. Additionally, data augmentation can help address issues of data scarcity and improve model robustness. Hybrid models that leverage both traditional regression methods and modern machine learning algorithms like LSTM may provide more versatile predictive capabilities. Real-time predictions could be a valuable extension, especially in applications such as finance and healthcare. The development of interpretable models is essential for understanding prediction rationales, and considerations of fairness, ethics, and transparency should guide future research in this field. Finally, exploring the practical aspects of model deployment, scalability, and integration into existing systems will be crucial for real-world applications, as well as extending these predictive techniques to various domains and industries.

## VII. REFERENCES

[1] G. Horng, H. Wu, Z. Wang, and G.-J. Jong, "A prediction method for voltage and lifetime of lead–acid battery by using machine learning," IEEE Transactions on Industry Applications, vol. 38, no. 6, pp. 1679-1686, Nov./Dec. 2002.

[2] Xiaodong Zhang, Shuang Yang, Bo Li, and Xingwei Li. A hybrid fault diagnosis approach for lithium-ion batteries based on a dynamic particle swarm optimization algorithm and an improved extreme learning machine. IEEE Access, 9:23696–23707, 2021.

[3] Bo Chen, Zhen Liu, Xiaojian Hu, Xinyue Wang, Hang Xiang, and Xuewei Zhang. Design and implementation of battery management system for electric vehicle application. Journal of Power Sources, 481:228800, 2021.

[4] Mengdie Tan, Wei Wang, Yue Yu, Guoqing Xu, and Jing Sun. Optimal design of battery management systems for electric vehicles considering battery performance characteristics. IEEE Transactions on Industrial Electronics, 65(7):5891–5901, 2018.

[5] Lijuan Zhang, Xingyu Jiang, Yi Huang, and Zhifeng Zou. Design of battery management system based on embedded system. Journal of Physics: Conference Series, 1357(3):032097, 2019.

[6] X. Zhang, H. Li, X. Wang, and G. Li. Journal of Energy Storage, 34:102335, 2021.

[7] Yong Li, Chao Wang, Haifeng Chen, and Xiaodong Lu. A review of model-based lithium-ion battery health estimation methods. Energies, 14(1):154, 2021.

[8] Xiaosong Zhang, Honglai Gao, Xin Jin, Xiaosong Guo, and Rui Zhang. Fault diagnosis of lithium-ion battery using wavelet transform and support vector machine. Journal of Power Sources, 239:680–687, 2013.

[9] Tao Yang, Guoqiang Liu, Chuan Li, and Hongwen He. Battery health prognosis based on a gaussian mixture model and support vector regression. IEEE Transactions on Vehicular Technology, 66(10):9028–9037, 2017.

[10] Mojtaba Alipour, Muhammad Khalid, Faisal Mohd-Yasin, Saad Mekhilef, and Mehdi Seyedmahmoudian. A review of model predictive control applied to energy management and building systems. Journal of Building Engineering, 12:235–253, 2017.

[11] Pierluigi Pisu and Hamid Reza Karimi. Model predictive control of battery systems for electric and hybrid vehicles. IEEE Transactions on Control Systems Technology, 20(3):696–704, 2012.

[12] Xiaojun Zhang, Binggang Cao, Jiaxin Li, Hui Wang, Rui Zhang, and Xiaohua Lu. A model predictive control strategy for lithium-ion battery thermal management in electric vehicles. Journal of Power Sources, 438:227031, 2019.

[13] Md Asif Uddin, John R Wagner, and Charles R Sullivan. A review of lithium-ion battery state-of-charge estimation and management system in electric vehicle applications: Challenges and recommendations. Renewable and Sustainable Energy Reviews, 78:834–854, 2017.

[14] Y. Li, M. Xie, and J. Wang. State-of-charge estimation for electric vehicles using neural network approach. Journal of Power Sources, 223:62–71, 2013.

[15] Yu Zhang, Huei Peng, and Yannian Yang. A review on state-of-charge estimation techniques for electric vehicle batteries. Renewable and Sustainable Energy Reviews, 78:438–450, 2017.

[16] Madeleine Ecker, Timo Kärger, and Andreas Jossen. Energy management of lithium-ion batteries: Diagnosis and prognosis. Journal

of Power Sources, 254:1–11, 2014.
[17] Shuliang Wen, Hongwen He, Xiaosong Zhang, Xiaowei Wang, and Yaoyao He. Energy management for electric buses based on a real-time battery model. IEEE Transactions on Transportation Electrification, 4(4):862–872, 2018.
[18] H. Li, Y. Yang, and C. Zou. Hierarchical control strategy for the integration of battery management system and vehicle control system in electric vehicles. Journal of Power Sources, 484:229242, 2021.
[19] Jie Cheng, Yuxin Zhang, Zhe Li, Jian Zhang, and Yanhong Li. Integrated energy management system for battery management system, charging infrastructure and power grid. Journal of Energy Storage, 30:101551, 2020.
[20] Hui He, Yucheng Li, and Jianqiu Li. Model-based control strategy for battery management system of electric vehicle. Energy Conversion and Management, 115:126–136, 2016.
[21] Fengchun Sun, Qiang Wang, and Haitao Yu. Electric vehicle battery management system: A literature review. Applied Energy, 179:413–427, 2016.
[22] Yu Duan, Bin Liu, and Zhuomin Zhang. A review on battery thermal management in electric vehicle application. Journal of Power Sources, 258:20–36, 2014.
[23] Zhe Huo, Dianbo Du, Feng Gao, Fengchun Sun, and Longya Xu. Review of battery thermal management systems for electric vehicle applications. Renewable and Sustainable Energy Reviews, 82:1395–1409, 2018.
[24] Y. Wang, H. He, W. Zhang, and J. Li. Battery thermal management in electric vehicles. IEEE Transactions on Vehicular Technology, 68(11):10510–10522, 2019.
[25] Bin Chen, Chao-Yang Wang, Zhonghua Wu, and Jianbo Zhang. Battery thermal management system design and analysis for electric vehicles. Energies, 8(8):8381–8394, 2015.
[26] Kyungho Kim, Seonho Park, Minhyeok Lee, and Soo-Kyun Kim. State estimation of lithium-ion batteries using an adaptive extended kalman filter. Energies, 10(12):2131, 2017.